\documentclass[journal]{IEEEtran}
\usepackage[font=small,labelsep=colon]{caption}
\usepackage{booktabs}

\usepackage{amssymb}
\usepackage{amsmath}

\usepackage{enumitem}
\ifCLASSINFOpdf
  \usepackage[pdftex]{graphicx}
\else
\fi
\begin{document}
%
\title{EfficientVITON: An Efficient Virtual Try-On Model using Optimized Diffusion Process}
%
%
%

\author{\IEEEauthorblockN{Mostafa Atef$^*$, Mariam Ayman$^*$, Ahmed Rashed$^*$, Ashrakat Saeed$^*$, Abdelrahman Saeed$^*$, Ahmed Fares$^*$}
\IEEEauthorblockA{$^*$Department of Computer Science and Engineering, Egypt-Japan University of Science and Technology, Alexandria, Egypt.}
}

\maketitle

\begin{abstract}
Wouldn't it be much more convenient for everybody to try on clothes by only looking into a mirror? The answer to that problem is  “virtual try-on," enabling users to digitally experiment with outfits. The core challenge lies in realistic image-to-image translation, where clothing must fit diverse human forms, poses, and figures. Early methods, which used 2D transformations, offered speed, but image quality was often disappointing and lacked the nuance of deep learning. Though GAN-based techniques enhanced realism, their dependence on paired data proved limiting. More adaptable methods offered great visuals but demanded significant computing power and time. Recent advances in diffusion models have shown promise for high-fidelity translation, yet the current crop of virtual try-on tools still struggle with detail loss and warping issues. To tackle these challenges, this paper proposes EfficientVITON: a new virtual try-on system leveraging the impressive pre-trained Stable Diffusion model for better images and deployment feasibility. The system includes a spatial encoder to maintain clothing's finer details and zero cross-attention blocks to capture the subtleties of how clothes fit a human body. Input images are carefully prepared, and the diffusion process has been tweaked to significantly cut generation time without image quality loss. The training process involves two distinct stages of fine-tuning, carefully incorporating a balance of loss functions to ensure both accurate try-on results and high-quality visuals. Rigorous testing on the VITON-HD dataset, supplemented with real-world examples, has demonstrated that EfficientVITON achieves state-of-the-art results. 
\end{abstract}


%
\IEEEpeerreviewmaketitle

\section{Introduction}
The rapid advancement of artificial intelligence (AI) and computer vision has revolutionized various industries, notably fashion and retail, with the emergence of virtual try-on technology.  This technology allows users to digitally "try on" garments using their own images, offering a personalized and engaging shopping experience \cite{vitonhd,LAION,stableviton}. Virtual try-on addresses key challenges in the fashion industry, including high return rates due to fit uncertainty, sustainability concerns related to excessive inventory and returns, and the need for enhanced customer engagement.

Despite the transformative potential of virtual try-on, creating effective systems presents significant technical hurdles. Traditional methods, relying on paired datasets and external warping networks, often lack generalizability to diverse body poses, complex backgrounds, and varying clothing styles \cite{hrVITON}.  Moreover, preserving intricate garment details and achieving real-time performance remain challenging due to computational overhead.

Recent progress in pre-trained diffusion models like Stable Diffusion offers promising avenues for high-quality image synthesis \cite{Rombach2021HighResolutionIS}.  However, adapting these models for virtual try-on requires addressing challenges in semantic alignment between clothing and body features, as well as improving processing speed without sacrificing image quality \cite{vitonhd,parserfreevton,highresvton}.

EfficientVITON tackles these challenges by combining the strengths of diffusion models with innovative architectural enhancements. We introduce an optimized diffusion process using non-uniform timesteps for improved computational efficiency. Furthermore, a spatial encoder and zero cross-attention blocks are incorporated to maintain garment details and achieve accurate alignment \cite{conditionalcontrol}.  These advancements allow EfficientVITON to handle diverse poses, backgrounds, and clothing styles while preserving textures and intricate features, leading to more realistic and efficient virtual try-on experiences.

The versatility of virtual try-on extends its applicability to various domains, including e-commerce integration for reduced return rates and improved customer satisfaction, physical retail stores for enhanced shopping experiences, personalized styling and fashion exploration, sustainable fashion practices, and entertainment and media applications. EfficientVITON's focus on speed and accuracy makes it particularly well-suited for real-time applications, furthering its potential for widespread adoption.

The sections of the paper are organized as follows:
\begin{enumerate}
    \item \textbf{Introduction:} Presents the motivation, problem statement, applications, and contributions of the work.
    \item \textbf{Related Work:} Reviews the state-of-the-art virtual try-on systems and highlights their limitations.
    \item \textbf{Methodology:} Describes the architecture of EfficientVITON, including the use of Stable Diffusion, zero cross-attention blocks, and non-uniform diffusion steps.
    \item \textbf{Experimental Results:} Presents quantitative and qualitative evaluations, comparing EfficientVITON with existing methods.
    \item \textbf{Conclusion} Summarizes the contributions and outlines potential directions for future research.
\end{enumerate}

\section{Related Work} 
Virtual try-on systems have become a cornerstone of contemporary e-commerce, allowing users to see how clothing items would look on them without the need for physical trials. Early approaches relied on 2D image warping and stitching techniques, which often fell short when handling complex poses, textures, or lighting variations. The emergence of deep learning, particularly generative models, has transformed this field by enabling more realistic and scalable solutions. In recent years, advances in diffusion models have further pushed the limits of image synthesis, delivering exceptional quality and stability in virtual try-on applications \cite{tryondiffusion}\cite{stableviton}.

Also, the evolution of Generative Models in Virtual Try-On Generative models, particularly Generative Adversarial Networks (GANs), have become a cornerstone of virtual try-on systems due to their ability to generate highly realistic images. Despite their popularity, GANs face several challenges, including mode collapse, training instability, and difficulties in preserving fine details in high-resolution outputs. As an alternative, diffusion models have gained traction by employing iterative denoising processes to produce high-quality images with greater stability. Recent advancements have showcased the potential of diffusion models in addressing fashion-specific challenges, including their application in virtual try-on tasks \cite{DCIVITON}\cite{ladivton}.

Over the past five years, virtual try-on systems have advanced significantly as researchers explore new models and techniques to overcome critical challenges, such as pose alignment, texture preservation, and image realism. Below is an overview of the key approaches, their contributions, and their limitations:
    \begin{enumerate}
        \item \textbf{Geometric Warping-Based Models}
        \par Early virtual try-on systems heavily relied on geometric warping techniques to align clothing items with the target body pose.
        These methods typically employed 2D image transformations, such as Thin Plate Spline (TPS), to warp garments onto the user’s image. For instance, VITON \cite{han2018viton} introduced a TPS-based warping approach that enhanced alignment but struggled to preserve fine details and textures. While computationally efficient, these methods often produced unrealistic results when dealing with complex poses, occlusions, or fabric deformations.\\
        \textbf{Strengths:} Low computational cost and straightforward implementation. \\
        \textbf{Weaknesses:} Poor handling of complex poses, textures, and overall realism.
        \item \textbf{GAN-Based Models}
In addition, the advent of GANs revolutionized virtual try-on systems by enabling the realistic synthesis of garments on target poses. These methods generally operate in two stages: first, a warping module aligns the clothing with the body, followed by a refinement module that generates the final output. For example, VITON \cite{han2018viton} and CP-VTON \cite{wang2018characteristic} are GAN-based frameworks that combined pose estimation with texture preservation to achieve state-of-the-art results. However, GANs often face issues such as training instability, visible artifacts, and challenges in preserving fine details.
        
        \textbf{Strengths:} High-quality realism and the ability to handle intricate textures. \\
        \textbf{Weaknesses:} Unstable training, artifact generation, and difficulty preserving details.
        \item \textbf{Attention-Based Models}
        \par To address GANs’ limitations, attention mechanisms were introduced to enhance alignment and texture preservation. These models use attention maps to focus on relevant regions of the clothing and the body, enabling more precise alignment and detailed output. For example, CP-VTON+ \cite{issenhuth2020parserfree} developed an attention-based try-on network that excelled at managing complex patterns and textures. However, attention-based models tend to be computationally intensive and require large datasets for training.
        
        \textbf{Strengths:} Better alignment and improved texture detail. \\
        \textbf{Weaknesses:} High computational cost and heavy data requirements.
        
        \item \textbf{Diffusion Models}
Recently, diffusion models have emerged as a powerful alternative to GANs in virtual try-on applications. Unlike GANs, these models iteratively denoise images, which enhances stability and output quality. They have proven highly effective at managing complex textur \cite{tryondiffusion} and Stable-VITON \cite{stableviton} are diffusion-based models that demonstrate exceptional image quality and stability. However, their iterative nature makes them computationally expensive, which poses challenges for real-time applications.
        
        \textbf{Strengths:} Exceptional image quality, stability, and detail preservation. \\
        \textbf{Weaknesses:} High computational demands and slow inference speeds.

        \item \textbf{Hybrid Models}
        \par Hybrid models combine the strengths of multiple approaches, such as geometric warping, GANs, and attention mechanisms, to achieve better alignment, realism, and efficiency. For instance, HR-VITON \cite{hrVITON} proposed a hybrid framework that integrates geometric warping for initial alignment with GAN-based refinement for final synthesis. While effective, hybrid models often require complex architectures and significant tuning.
        
        \textbf{Strengths:} Balanced performance by leveraging multiple techniques. \\
        \textbf{Weaknesses:} Complex architecture and intensive tuning requirements.
    
        \item \textbf{Multi-Modal Models}
        \par Advancements in multi-modal inputs, such as text descriptions and sketches, have enhanced user interaction and personalization in virtual try-on systems. For example, Text2Cloth \cite{ren2021clothtransformer} developed a system allowing users to describe clothing in natural language, which the model then synthesizes onto their image. Similarly, Sketch2TryOn \cite{xie2021patchgan} introduced a sketch-based interface for designing and visualizing custom clothing in real-time. While engaging, these methods require additional preprocessing and are computationally expensive.
        
        \textbf{Strengths:} Greater personalization and user engagement. \\
        \textbf{Weaknesses:} High computational cost and preprocessing demands.
    
    \end{enumerate}
    
Diffusion-based methods currently lead the field, generating high-resolution, photorealistic images that preserve intricate clothing details. Recent advancements include integrating pose estimation, attention mechanisms for alignment, and multi-modal inputs for personalization. However, challenges such as adapting to diverse body shapes, managing occlusions, and improving real-time performance remain unresolved.

Despite significant progress, several gaps persist in virtual try-on research. Current methods often focus solely on static images, underutilizing temporal data for video-based try-on systems \cite{densepose}. Personalization, such as accommodating user preferences or diverse body types, remains limited. The computational cost of diffusion models also restricts their use in real-time applications. Furthermore, there is a lack of standardized evaluation metrics tailored to virtual try-on systems, complicating objective comparisons between methods \cite{issenhuth2020parserfree}.

While existing systems have improved realism and usability, many struggle to generalize across diverse datasets and real-world scenarios. For instance, GAN-based methods often generate artifacts in complex scenarios, while diffusion models, despite their stability, remain computationally expensive \cite{tryondiffusion}\cite{stableviton}. Additionally, most systems rely on paired datasets, limiting scalability.

The research proposes a diffusion model framework to improve virtual try-on experiences, utilizing non-uniform time steps, temporal coherence, and personalization to reduce computational demands and enhance image quality.

\section{Methodology} 
This section details the EfficientVITON framework for virtual try-on, leveraging a pre-trained diffusion model and incorporating optimizations for efficiency and enhanced clothing detail preservation.

\subsection{Data \& Preprocessing}

\begin{figure}[h]
  \centering
  \includegraphics[width=0.5\columnwidth]{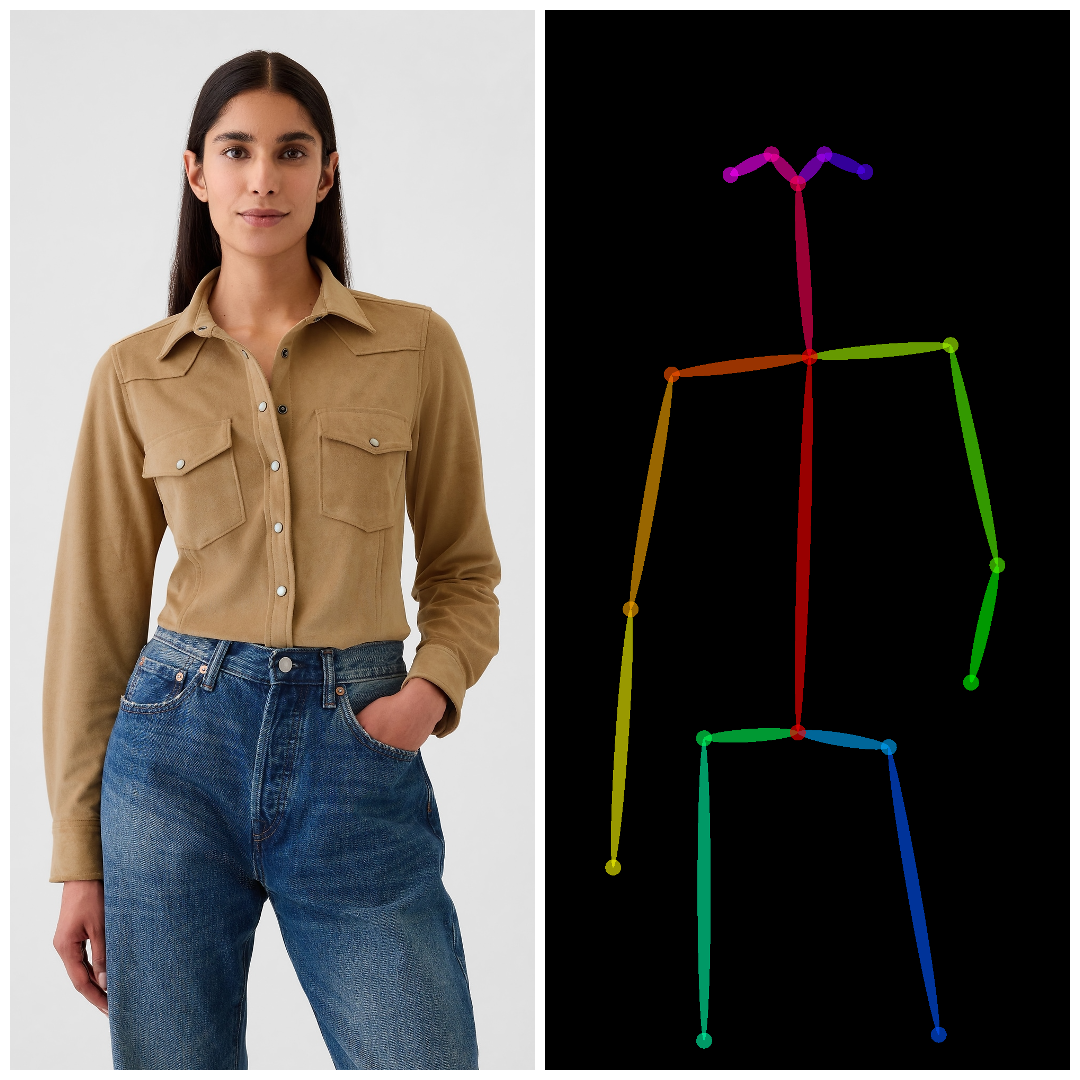}
  \caption{OpenPose Output.}
  \label{fig:openpose_out}
\end{figure}

\begin{figure}[h]
  \centering
  \includegraphics[width=0.5\columnwidth]{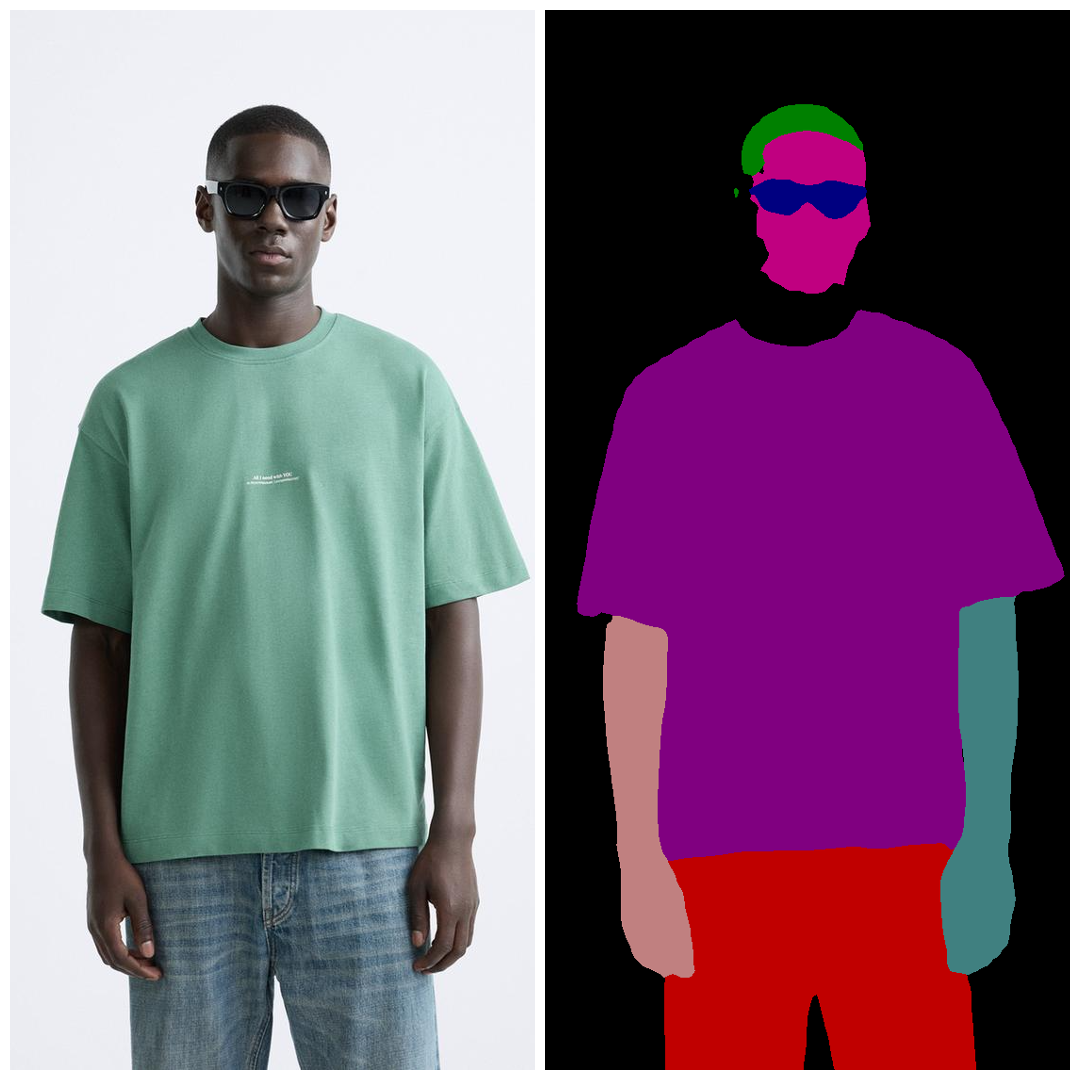}
  \caption{LIP Parsing Output.}
  \label{fig:lip_out}
\end{figure}

\begin{figure}[h]
  \centering
  \includegraphics[width=0.5\columnwidth]{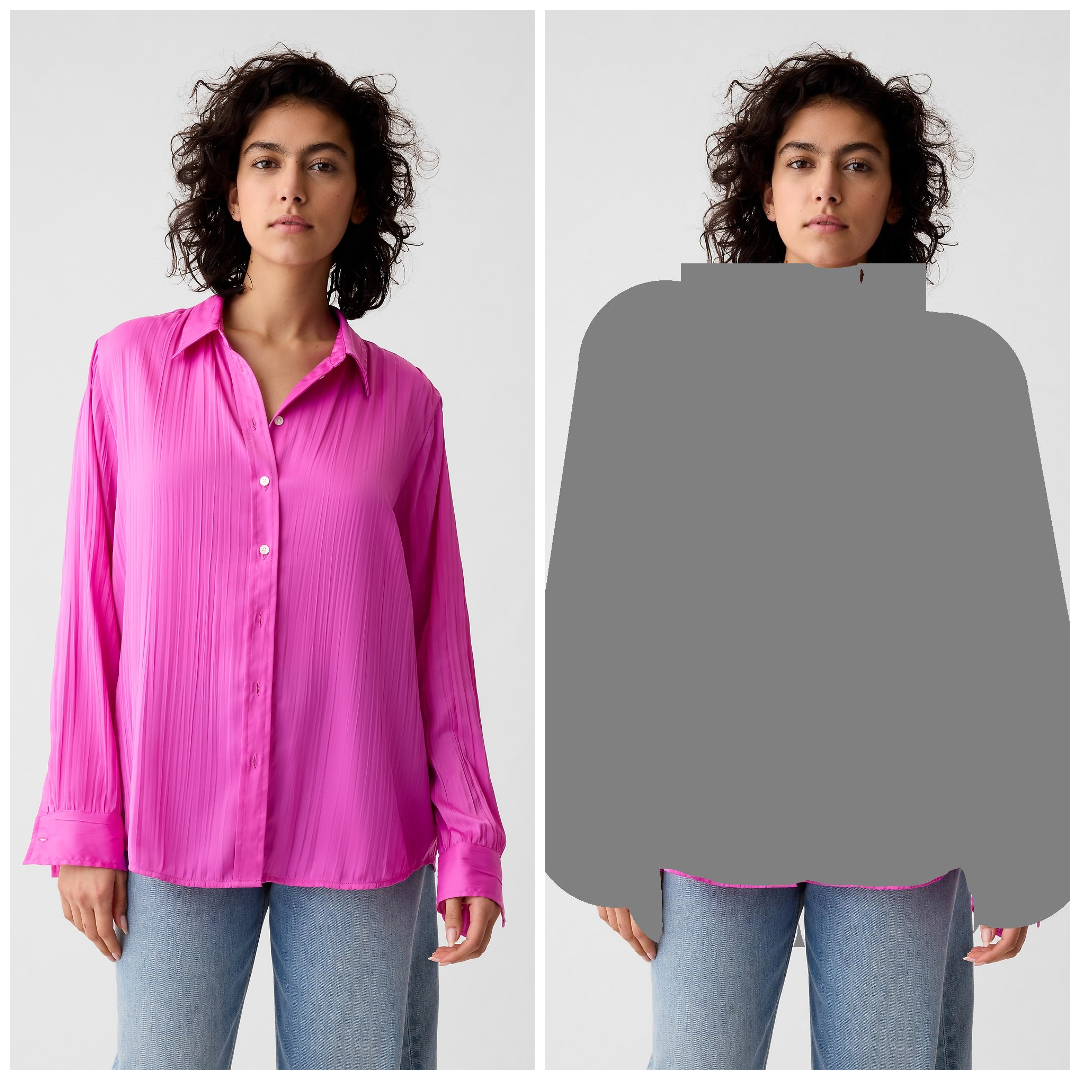}
  \caption{Agnostic Image Output.}
  \label{fig:agnostic_out}
\end{figure}

\begin{figure}[h]
  \centering
  \includegraphics[width=0.5\columnwidth]{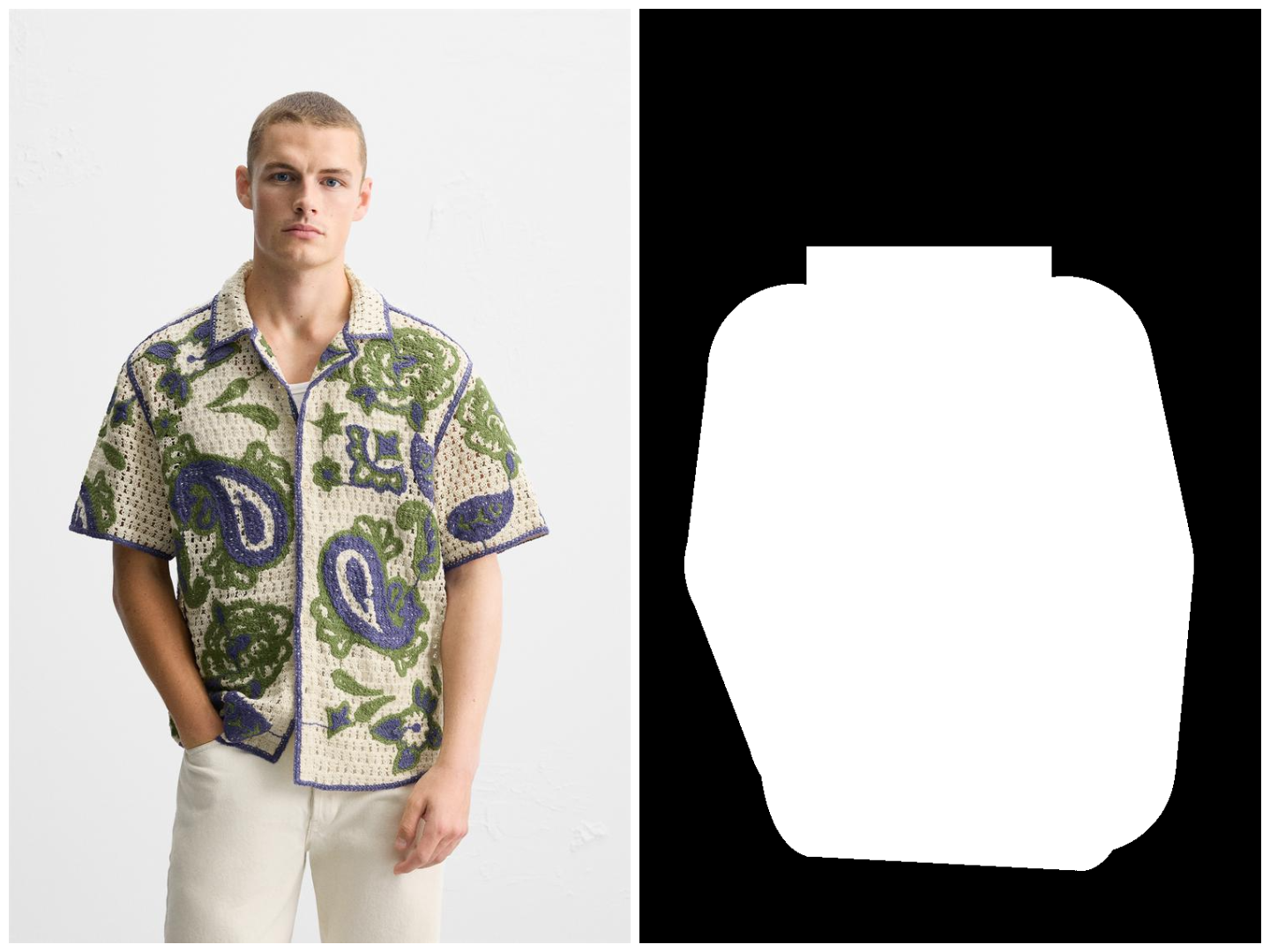}
  \caption{Agnostic Mask Output.}
  \label{fig:agnostic_mask_out}
\end{figure}

\begin{figure}[h]
  \centering
  \includegraphics[width=0.5\columnwidth]{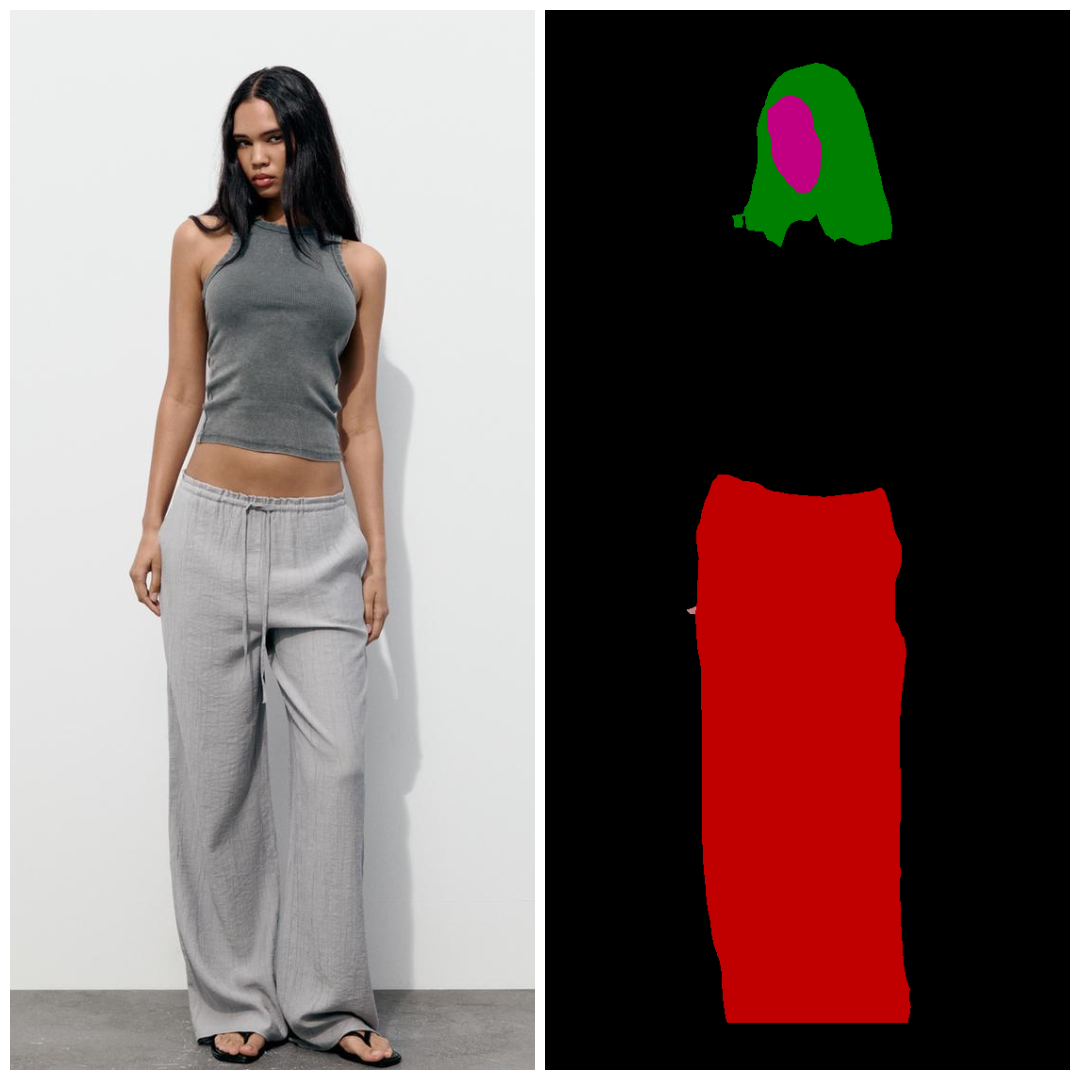}
  \caption{Parse Agnostic Image Output.}
  \label{fig:parse_agnostic_out}
\end{figure}

\begin{figure}[h]
  \centering
  \includegraphics[width=0.5\columnwidth]{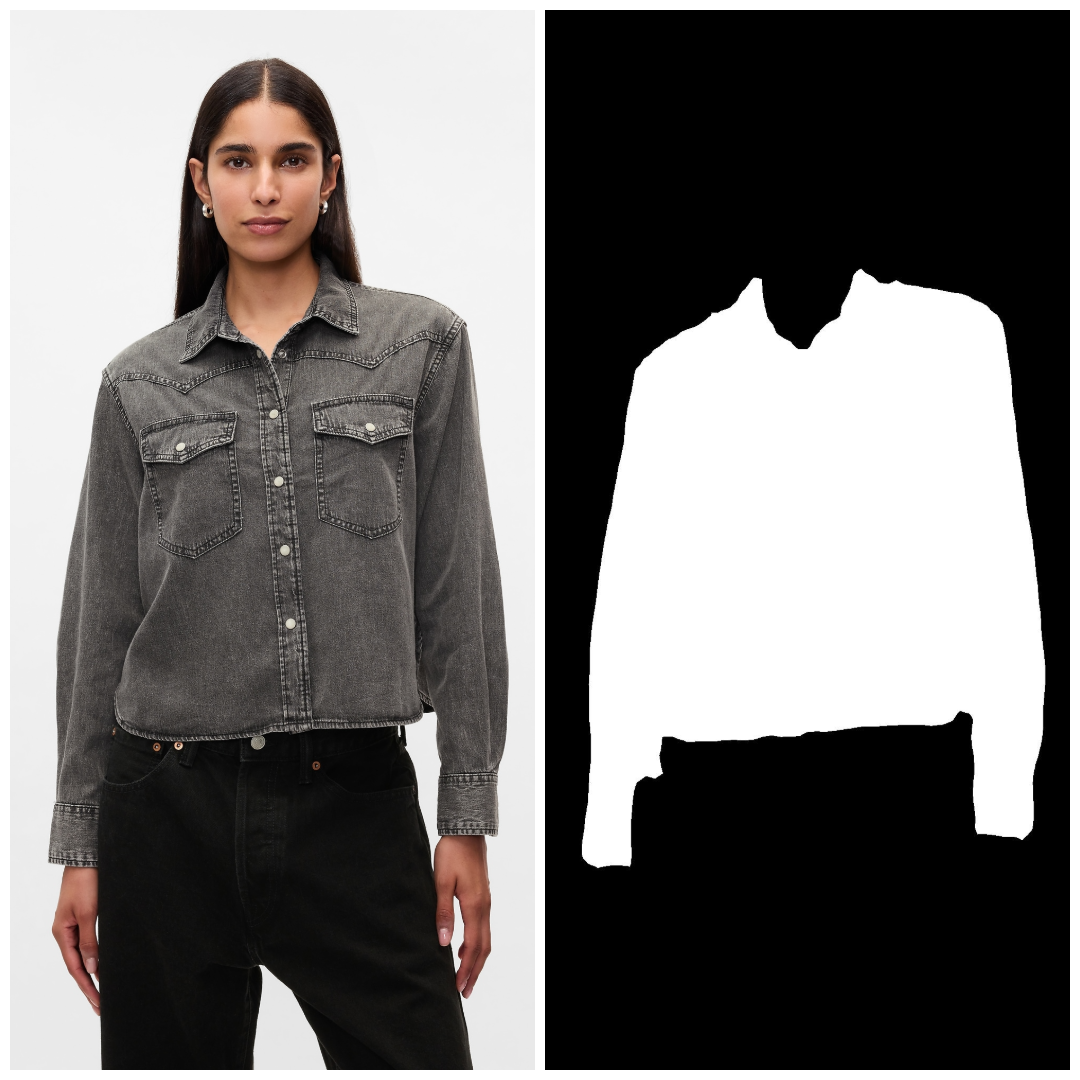}
  \caption{Ground Truth Warp Mask Output.}
  \label{fig:gt_warp_out}
\end{figure}

\begin{figure}[h]
  \centering
  \includegraphics[width=0.5\columnwidth]{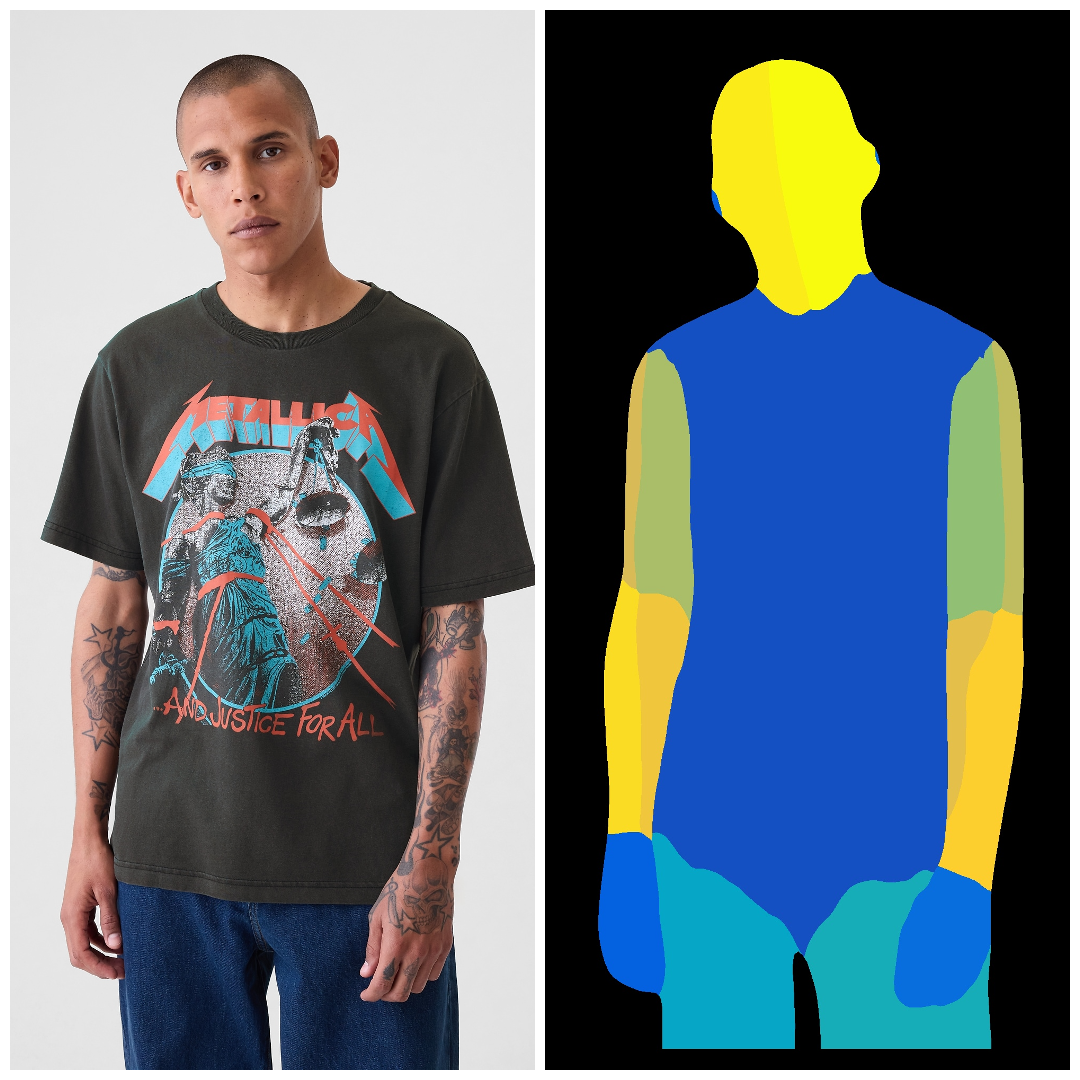}
  \caption{DensePose Output.}
  \label{fig:densepose_out}
\end{figure}

\begin{figure}[h]
  \centering
  \includegraphics[width=0.5\columnwidth]{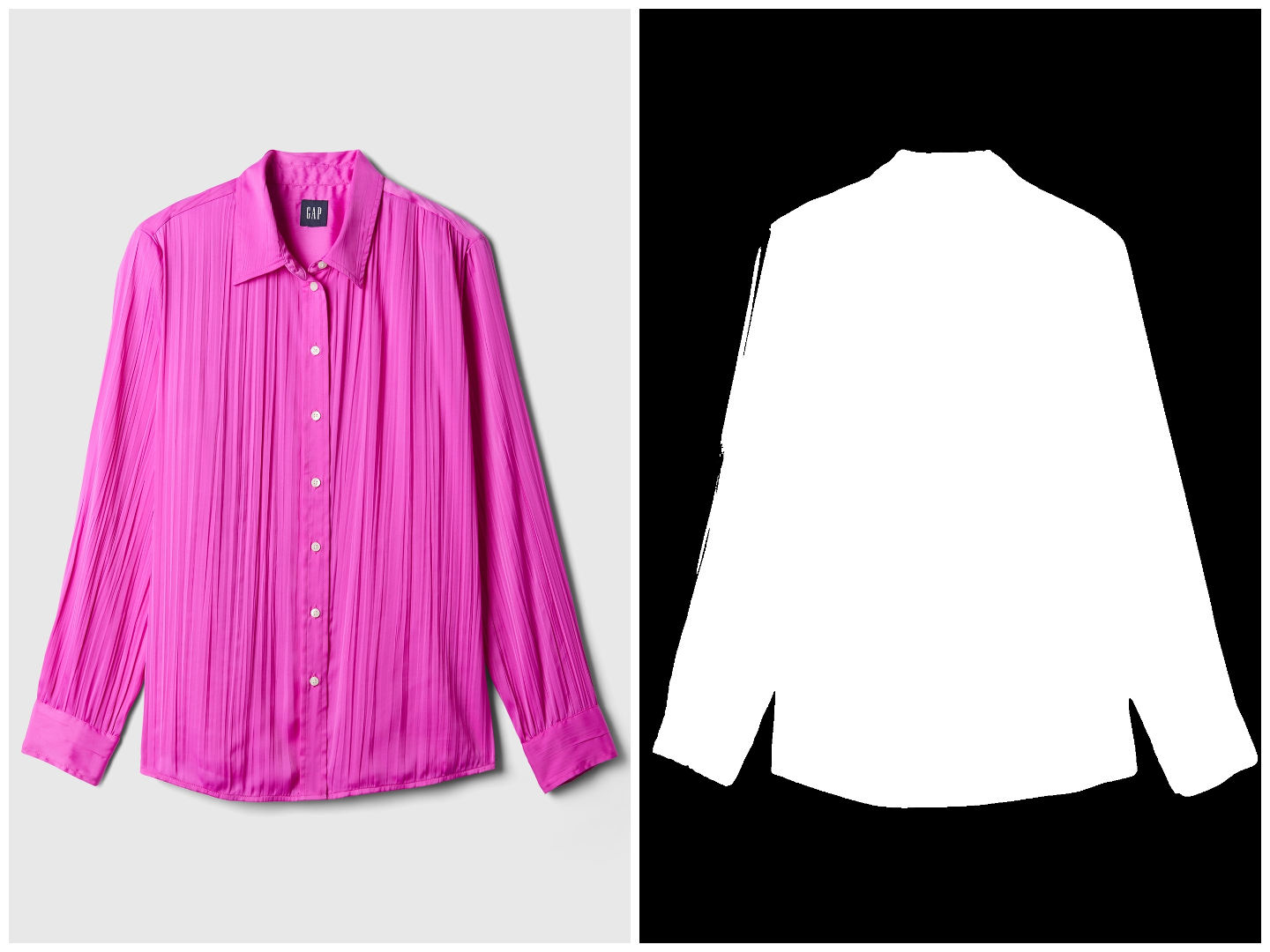}
  \caption{Cloth Mask Output.}
  \label{fig:cloth_mask_out}
\end{figure}

\begin{figure}[h]
  \centering
  \includegraphics[width=0.5\columnwidth]{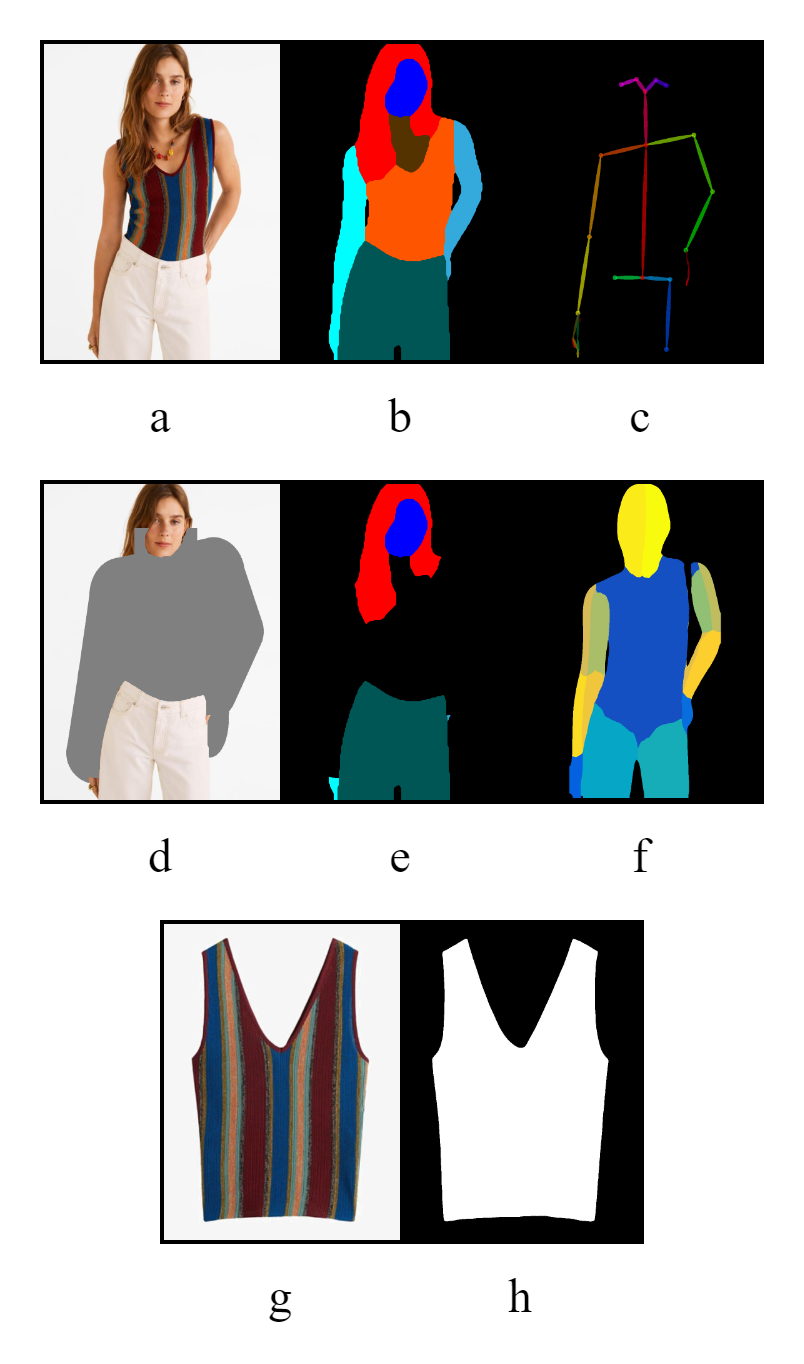}
  \caption{A sample from VITON-HD dataset. a) Normal Person Image. b) Human Parsing. c) OpenPose Pose Estimation. d) Agnostic Image. e) Parse Agnostic Image. f) DensePose Image. g) Unworn Cloth Image. h) Cloth Mask.}
  \label{fig:vitonhd}
\end{figure}

EfficientVITON utilizes a preprocessed dataset derived from VITON-HD \cite{vitonhd}, applying the following steps (Fig. \ref{fig:vitonhd}):

\begin{enumerate}[label=\textbf{\arabic*)}]
    \item \textbf{Pose Estimation:} OpenPose (Fig. \ref{fig:openpose_out}) extracts 25 keypoints to define body regions \cite{openpose1,openpose2,openpose3,openpose4}.
    \item \textbf{Human Parsing:} LIP Parsing \cite{LIP} segments the person image into 20 semantic parts (Fig. \ref{fig:lip_out}).
    \item \textbf{Agnostic Image:} A grey mask, based on pose and parsing, covers the original clothing area (Fig. \ref{fig:agnostic_out}).
    \item \textbf{Agnostic Mask:} A binary mask isolates the original clothing region (Fig. \ref{fig:agnostic_mask_out}).
    \item \textbf{Parse Agnostic:} The agnostic mask area is removed from the parsed semantic map (Fig. \ref{fig:parse_agnostic_out}).
    \item \textbf{Ground Truth Warp Mask:}  A mask of the worn garment is extracted for training (Fig. \ref{fig:gt_warp_out}).
    \item \textbf{Dense Human Pose Estimation:} DensePose \cite{densepose} generates a UV map (Fig. \ref{fig:densepose_out}), subsequently encoded into a latent representation, providing detailed 3D body surface information.
    \item \textbf{Cloth Mask:}  A binary mask of the garment image is extracted (Fig. \ref{fig:cloth_mask_out}).
\end{enumerate}

\subsection{Stable Diffusion Architecture}

EfficientVITON builds upon the Stable Diffusion model \cite{Rombach2021HighResolutionIS}, leveraging its high-fidelity generation, latent space efficiency, and pre-trained knowledge of human and clothing features.  Stable Diffusion consists of a Variational Autoencoder (VAE) for latent space compression and reconstruction, a U-Net for denoising, and a diffusion process operating in the latent space.

While the quality of the output of Stable Diffusion is relatively high, it is not efficient enough. It uses a lot of memory resources and needs a lot of time to complete the diffusion process. Jiang et al. addressed the problem of an efficient diffusion process in their research \cite{fastddpm}, in which we took a similar approach to their solution to this problem. The main solution to this problem is to modify the timesteps required for the model to do the denoising process. Instead of a large number of uniformly distributed timesteps, we apply a non-uniform distribution of a small number of timesteps in the denoising process.

Instead of sampling \textit{\textbf{n}} steps uniformly from all possible timesteps, we sample from a smaller set of strategically chosen timesteps. This significantly reduces the time needed for the diffusion process without sacrificing the quality of the output image, as the model will now learn the most significant timesteps in the denoising process and try to take more impulsive steps towards the final goal. The non-uniform distribution allows the model to take different amounts of denoising steps according to the position of the timestep. Fig. \ref{fig:efficient} shows a description of the non-uniform denoising steps which allows the model to be more efficient.

\subsection{EfficientVITON Architecture}

\begin{figure*}[h]
  \centering
  \includegraphics[width=0.8\textwidth]{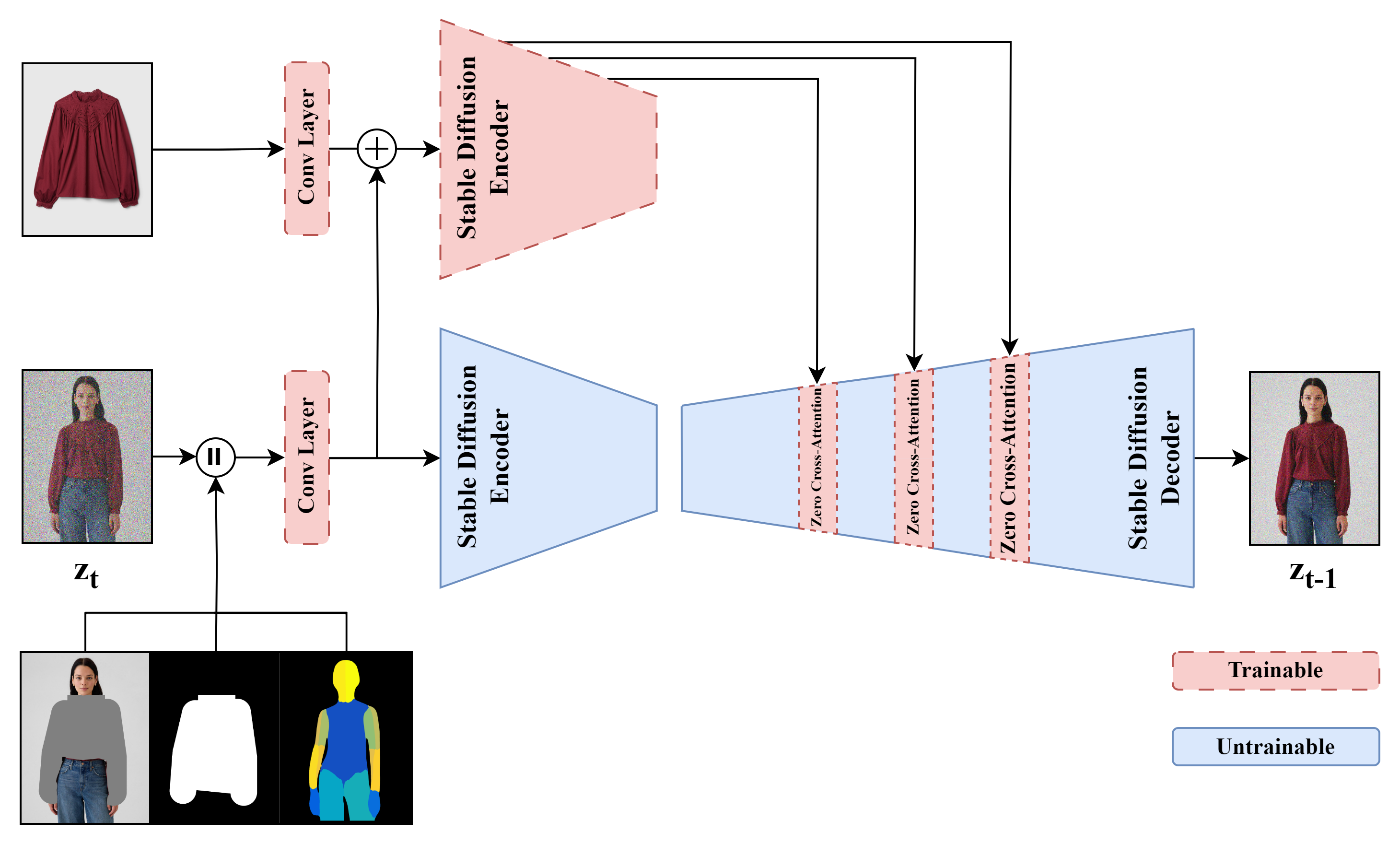}
  \caption{EfficientVITON Architecture.}
  \label{fig:vton_arch}
\end{figure*}

EfficientVITON Architecture (Fig. \ref{fig:vton_arch}) integrates a spatial encoder and zero cross-attention blocks with the Stable Diffusion core to perform virtual try-on.

\noindent\textbf{Inputs:} The model receives the preprocessed clothing image ($x_c$), agnostic map ($x_a$), agnostic mask ($x_{ma}$), and latent dense pose ($x_p$).

\noindent\textbf{Spatial Encoder:}  This encoder, initialized with U-Net weights, processes the clothing image ($x_c$) and extracts multi-resolution feature maps, capturing fine-grained clothing details.  These features are then used as key (K) and value (V) inputs to the zero cross-attention blocks.

\begin{figure}[h]
  \centering
  \includegraphics[width=0.5\columnwidth]{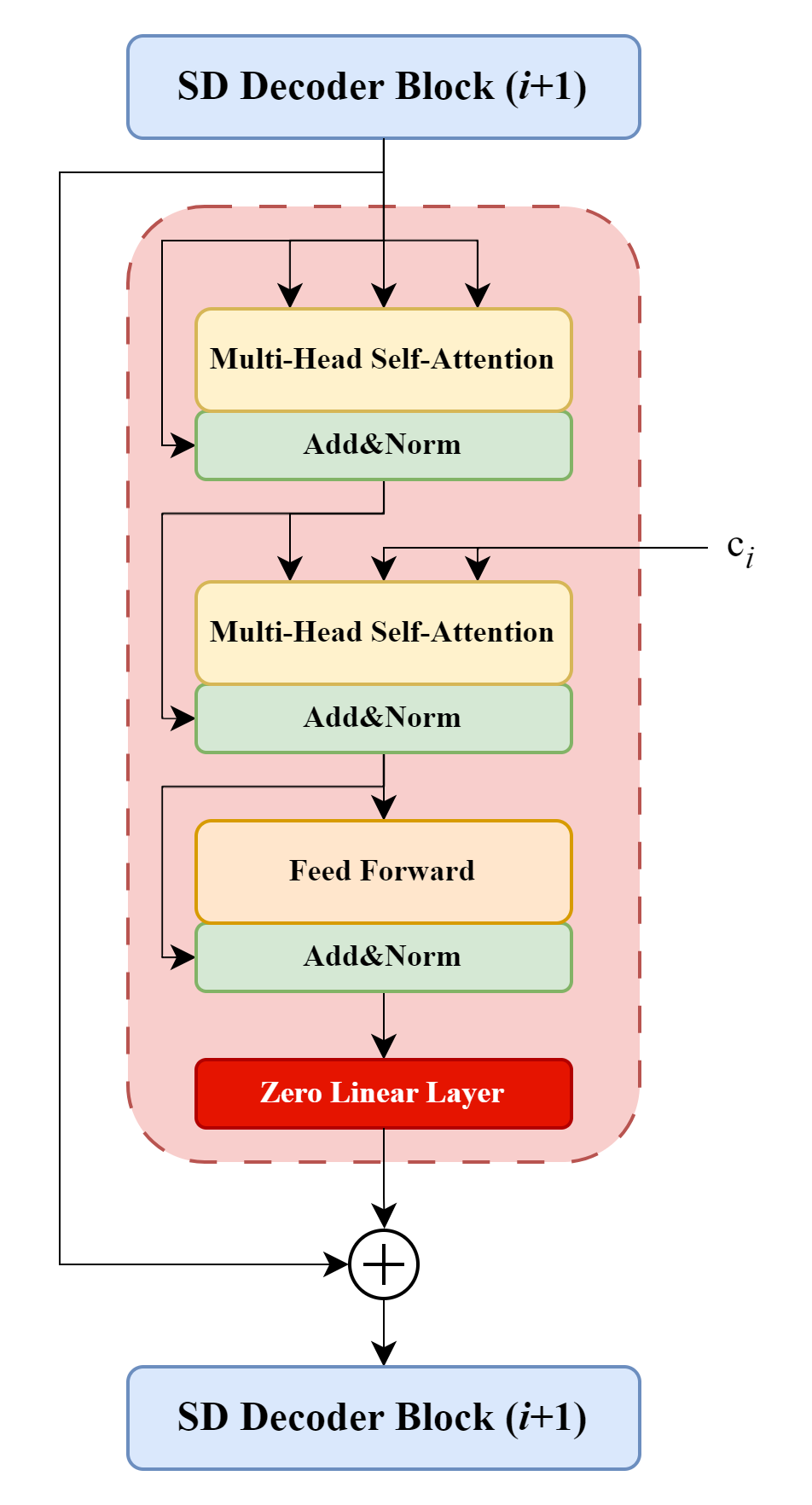}
  \caption{Zero Cross-Attention Block.}
  \label{fig:zero_attention}
\end{figure}

\noindent\textbf{Zero Cross-Attention Blocks} (Fig. \ref{fig:zero_attention}): Integrated within the U-Net decoder, these blocks learn the semantic correspondence between the clothing and the human body. The blocks use the spatial encoder's clothing features (K, V) and U-Net decoder features (Q) to perform patch-wise warping in the latent space.  A linear layer with zero-initialized weights helps to reduce noise.

\subsection{Efficient Diffusion Process}

\begin{figure}[h]
  \centering
  \includegraphics[width=\columnwidth]{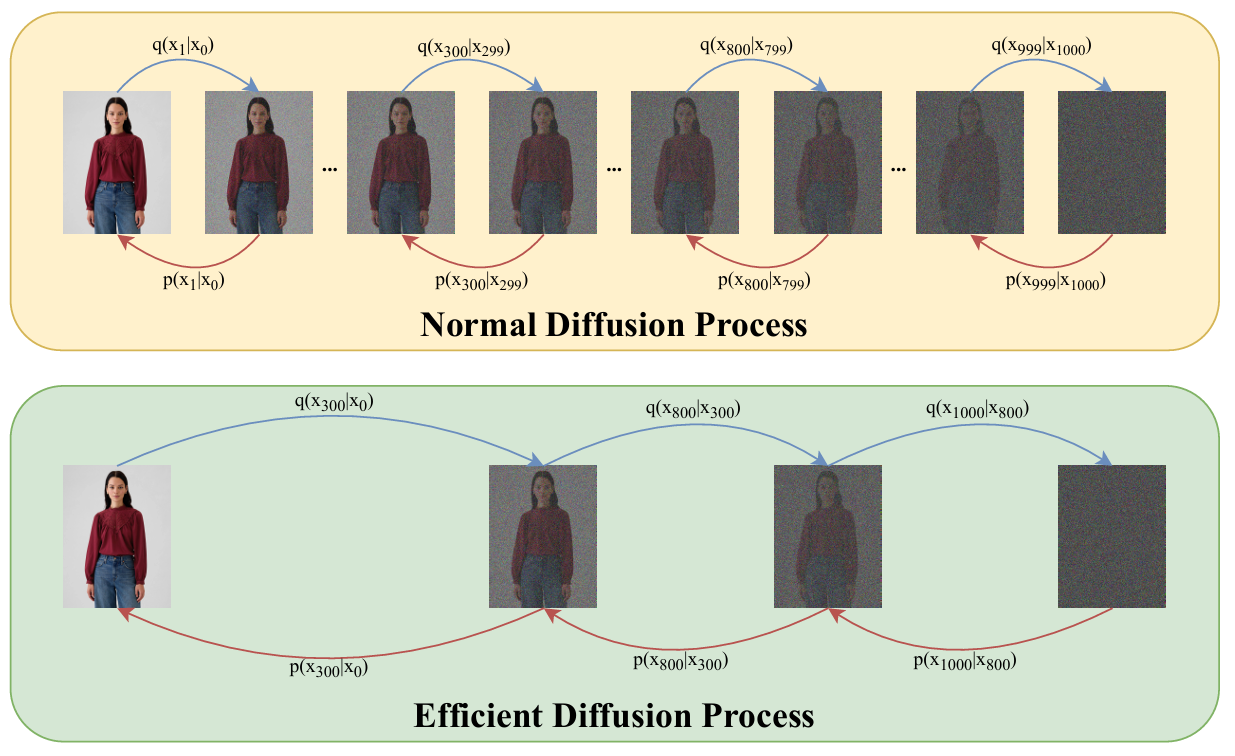}
  \caption{Efficient Diffusion vs. Standard Diffusion.}
  \label{fig:efficient}
\end{figure}

EfficientVITON optimizes the diffusion process by employing non-uniform timestep sampling \cite{fastddpm} (Fig. \ref{fig:efficient}). This concentrates denoising steps at critical timesteps, reducing computation while maintaining output quality.

\subsubsection{Loss Functions}

EfficientVITON combines the Stable Diffusion loss ($L_{LDM}$) with an Attention Total Variation Loss ($L_{ATV}$) to refine attention maps:

\begin{align}
    L_{LDM} &= \mathbb{E}_{\zeta, x_c, \epsilon, t} \lVert \epsilon - \epsilon_\theta(\zeta, t, T_\phi(x_c), E(x_c)) \rVert^2 \\
    L_{ATV} &= \lVert \nabla(FM) \rVert_1 \\
    L_{total} &= L_{LDM} + \lambda_{ATV} L_{ATV}
\end{align}

\subsubsection{Training and Inference}

Training is a two-stage process: (1) learning semantic correspondence with augmented inputs, and (2) refining attention maps with $L_{ATV}$.  Inference uses the PLMS sampler and, for paired evaluations, the RePaint \cite{RePaint} approach.

EfficientVITON's key contributions are: (1) end-to-end virtual try-on with a pre-trained diffusion model, (2) latent space semantic correspondence learning via zero cross-attention, (3) attention total variation loss and augmentation, and (4) an efficient diffusion process through non-uniform timestep sampling.

\section{Results} 
This section presents the qualitative, quantitative, and efficiency evaluation of EfficientVITON, demonstrating its superior performance in virtual try-on.

\subsection{Qualitative Results}

\begin{figure*}[h]
  \centering
  \includegraphics[width=0.850\textwidth]{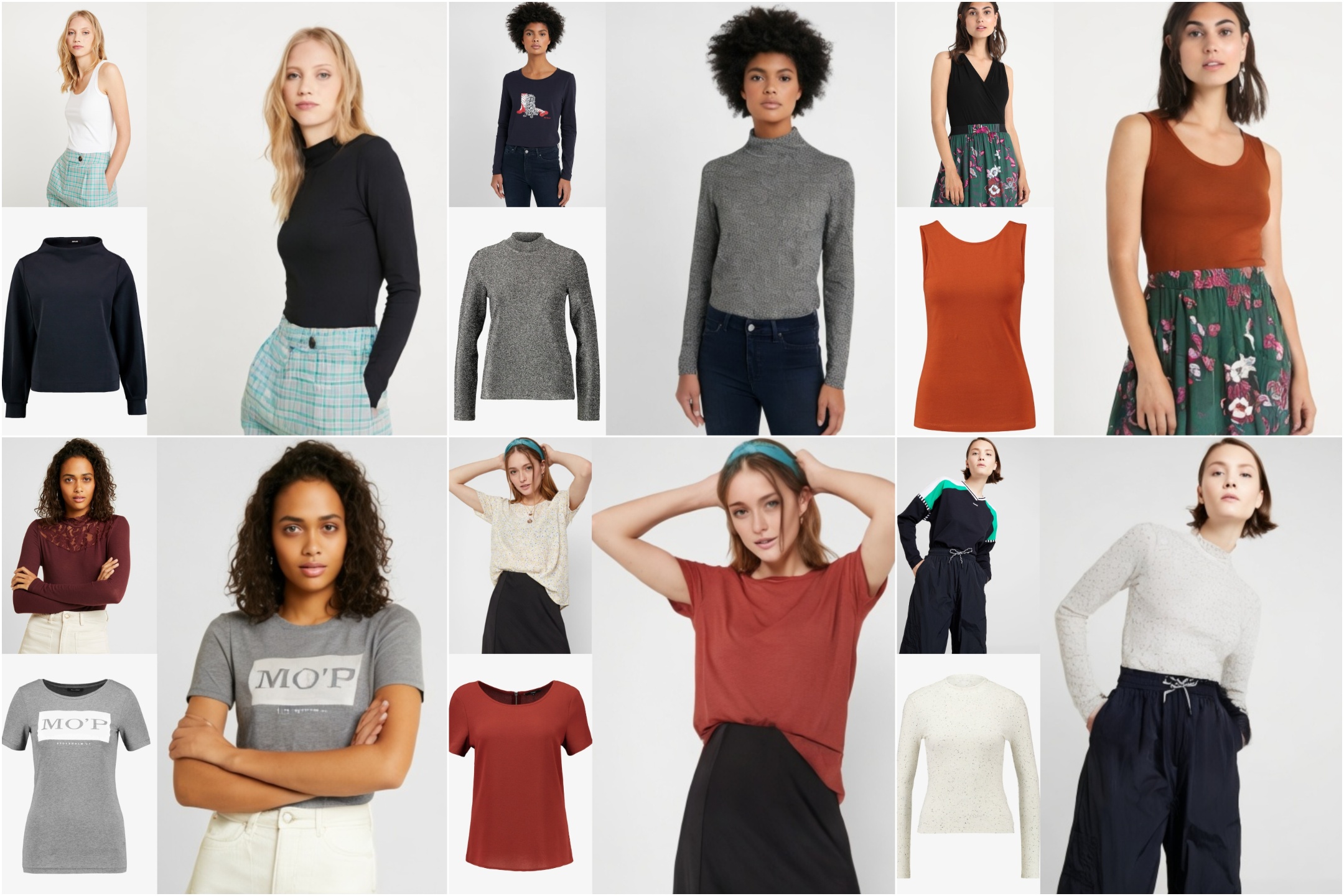}
  \caption{Qualitative Results on VITON-HD.}
  \label{fig:qualitative_results}
\end{figure*}

EfficientVITON generates visually realistic and accurate virtual try-on images (Fig. \ref{fig:qualitative_results}), effectively handling complex clothing patterns, text, logos, and varying body types and poses.  The model preserves intricate clothing details, maintains natural folds and shadows, and adapts seamlessly to different garment coverage levels, producing results visually indistinguishable from real photographs.

\subsection{Quantitative Results}

Quantitative evaluation using FID \cite{FID} and LPIPS \cite{LPIPS} (Table \ref{tab:comparison}) demonstrates EfficientVITON's superior performance.  The model achieves a lower FID score than state-of-the-art methods, indicating higher realism. While LPIPS scores are competitive, the RePaint \cite{RePaint} refinement further improves performance in paired settings by addressing potential reconstruction errors related to the agnostic map.  EfficientVITON's consistent performance across different experimental setups highlights its robustness and generalizability.

\setlength{\tabcolsep}{20pt}
\begin{table}[h!]
\centering
\begin{tabular}{lll} \toprule
Method & LPIPS  & FID \\ \midrule
VITON-HD \cite{vitonhd} & 0.117  & 12.117 \\
HR-VITON \cite{hrVITON} & 0.1045  & 11.265\\ 
LADI-VTON \cite{ladivton} & 0.0964 & 9.480 \\
Paint-by-Example  \cite{PaintbyExample} & 0.1428 & 11.939 \\
DCI-VTON \cite{DCIVITON} & 0.0804  & 8.754 \\
GP-VTON \cite{GP-VTON} & 0.088 & 9.072 \\\midrule
Ours & 0.0842 & 8.703  \\
Ours (RePaint) & 0.0762 & 8.433 \\\bottomrule
\end{tabular}
\caption{Quantitative Comparison (Lower is Better).}
\label{tab:comparison}
\end{table}

\subsection{Efficiency Results}

The non-uniform timestep sampling strategy significantly improves efficiency (Table \ref{tab:efficiency}), reducing both training and inference times. The inference time is reduced by 72.4\%, enabling near real-time performance, while training time is reduced by 45.3\%, facilitating faster model development and experimentation.  These improvements enhance practicality for real-world applications and resource-constrained environments.

\setlength{\tabcolsep}{20pt}
\begin{table}[h!]
\centering
\begin{tabular}{lll} \toprule
Method & Training time & Inference Time \\ \midrule
Before & 1570 h & 58 s \\
After & 859 h & 16 s \\ \bottomrule
\end{tabular}
\caption{Comparison of Training time and Inference time for our model before and After using our new approach of applying the non-uniform timestep distribution.}
\label{tab:efficiency}
\end{table}

\section{Conclusion}
This work introduces EfficientVITON, a novel virtual try-on framework that addresses the challenges of realism and efficiency. By incorporating non-uniform timestep sampling \cite{fastddpm} into a pre-trained diffusion model, EfficientVITON significantly reduces computational overhead (45.3\% reduction in training time and 72.4\% in inference time) without sacrificing visual fidelity. The inclusion of a spatial encoder and zero cross-attention blocks \cite{stableviton} further enhances realism by preserving fine clothing details and ensuring accurate alignment with the target body.

Quantitative results using FID \cite{FID} and LPIPS \cite{LPIPS} demonstrate EfficientVITON's superior performance compared to state-of-the-art GAN-based and diffusion-based virtual try-on methods. Qualitative results further confirm the model's ability to generate realistic and visually appealing try-on images across diverse poses, skin tones, and clothing styles.

EfficientVITON offers several key contributions: (1) an end-to-end virtual try-on framework based on a pre-trained diffusion model, eliminating the need for separate warping modules; (2) learning semantic correspondence directly in the latent space; (3) attention total variation loss and data augmentation for improved attention maps; and (4) enhanced efficiency through non-uniform timestep sampling.

This work has broader implications beyond virtual try-on. The optimized diffusion process using non-uniform timesteps can be applied to other image synthesis tasks. Moreover, EfficientVITON's efficiency makes it suitable for real-world applications in e-commerce, offering a personalized and engaging online shopping experience, potentially reducing return rates and promoting sustainability.  EfficientVITON represents a significant advancement in virtual try-on technology, bridging the gap between high-quality image synthesis and computational efficiency, and opening up new avenues for future research and development in generative modeling.


%

\ifCLASSOPTIONcaptionsoff
  \newpage
\fi



%

%

\bibliographystyle{IEEEtran}
\bibliography{ref}

\begin{thebibliography}{10}
\providecommand{\url}[1]{#1}
\csname url@samestyle\endcsname
\providecommand{\newblock}{\relax}
\providecommand{\bibinfo}[2]{#2}
\providecommand{\BIBentrySTDinterwordspacing}{\spaceskip=0pt\relax}
\providecommand{\BIBentryALTinterwordstretchfactor}{4}
\providecommand{\BIBentryALTinterwordspacing}{\spaceskip=\fontdimen2\font plus
\BIBentryALTinterwordstretchfactor\fontdimen3\font minus \fontdimen4\font\relax}
\providecommand{\BIBforeignlanguage}[2]{{%
\expandafter\ifx\csname l@#1\endcsname\relax
\typeout{** WARNING: IEEEtran.bst: No hyphenation pattern has been}%
\typeout{** loaded for the language `#1'. Using the pattern for}%
\typeout{** the default language instead.}%
\else
\language=\csname l@#1\endcsname
\fi
#2}}
\providecommand{\BIBdecl}{\relax}
\BIBdecl

\bibitem{vitonhd}
S.~Choi, S.~Park, M.~Lee, and J.~Choo, ``Viton-hd: High-resolution virtual try-on via misalignment-aware normalization,'' in \emph{Proceedings of the IEEE/CVF Conference on Computer Vision and Pattern Recognition}, 2021, pp. 14\,131--14\,140.

\bibitem{LAION}
C.~Schuhmann, R.~Beaumont, R.~Vencu, C.~Gordon, R.~Wightman, M.~Cherti, T.~Coombes, A.~Katta, C.~Mullis, M.~Wortsman, P.~Schramowski, S.~Kundurthy, K.~Crowson, L.~Schmidt, R.~Kaczmarczyk, and J.~Jitsev, ``Laion-5b: an open large-scale dataset for training next generation image-text models,'' in \emph{Proceedings of the 36th International Conference on Neural Information Processing Systems}, ser. NIPS '22.\hskip 1em plus 0.5em minus 0.4em\relax Red Hook, NY, USA: Curran Associates Inc., 2024.

\bibitem{stableviton}
J.~Kim, G.~Gu, M.~Park, S.~Park, and J.~Choo, ``Stableviton: Learning semantic correspondence with latent diffusion model for virtual try-on,'' in \emph{Proceedings of the IEEE/CVF Conference on Computer Vision and Pattern Recognition}, 2024, pp. 8176--8185.

\bibitem{hrVITON}
Q.~Lyu, Q.~Wang, and K.~Huang, ``High-resolution virtual try-on network with coarse-to-fine strategy,'' \emph{Journal of Physics: Conference Series}, vol. 1880, p. 012009, 04 2021.

\bibitem{Rombach2021HighResolutionIS}
\BIBentryALTinterwordspacing
R.~Rombach, A.~Blattmann, D.~Lorenz, P.~Esser, and B.~Ommer, ``High-resolution image synthesis with latent diffusion models,'' \emph{2022 IEEE/CVF Conference on Computer Vision and Pattern Recognition (CVPR)}, pp. 10\,674--10\,685, 2021. [Online]. Available: \url{https://api.semanticscholar.org/CorpusID:245335280}
\BIBentrySTDinterwordspacing

\bibitem{parserfreevton}
Y.~Ge, Y.~Song, R.~Zhang, C.~Ge, W.~Liu, and P.~Luo, ``Parser-free virtual try-on via distilling appearance flows,'' in \emph{Proceedings of the IEEE/CVF Conference on Computer Vision and Pattern Recognition (CVPR)}, 2021, pp. 8485--8493.

\bibitem{highresvton}
S.~Lee, G.~Gu, S.~Park, S.~Choi, and J.~Choo, ``High-resolution virtual try-on with misalignment and occlusion-handled conditions,'' in \emph{Proceedings of the European Conference on Computer Vision (ECCV)}.\hskip 1em plus 0.5em minus 0.4em\relax Springer, 2022, pp. 204--219.

\bibitem{conditionalcontrol}
L.~Zhang, A.~Rao, and M.~Agrawala, ``Adding conditional control to text-to-image diffusion models,'' in \emph{Proceedings of the IEEE/CVF International Conference on Computer Vision (ICCV)}, 2023, pp. 3836--3847.

\bibitem{tryondiffusion}
L.~Zhu, D.~Yang, T.~Zhu, F.~Reda, W.~Chan, C.~Saharia, M.~Norouzi, and I.~Kemelmacher-Shlizerman, ``Tryondiffusion: A tale of two unets,'' in \emph{Proceedings of the IEEE/CVF Conference on Computer Vision and Pattern Recognition (CVPR)}, June 2023, pp. 4606--4615.

\bibitem{DCIVITON}
J.~Gou, S.~Sun, J.~Zhang, J.~Si, C.~Qian, and L.~Zhang, ``Taming the power of diffusion models for high-quality virtual try-on with appearance flow,'' \emph{arXiv preprint arXiv:2308.06101}, 2023.

\bibitem{ladivton}
D.~Morelli, A.~Baldrati, G.~Cartella, M.~Cornia, M.~Bertini, and R.~Cucchiara, ``Ladi-vton: Latent diffusion textual-inversion enhanced virtual try-on,'' \emph{arXiv preprint arXiv:2305.13501}, 2023.

\bibitem{han2018viton}
X.~Han, Z.~Wu, Z.~Wu \emph{et~al.}, ``Viton: An image-based virtual try-on network,'' in \emph{Proceedings of the IEEE Conference on Computer Vision and Pattern Recognition (CVPR)}, 2018, pp. 7543--7552.

\bibitem{wang2018characteristic}
B.~Wang, H.~Zheng, X.~Liang \emph{et~al.}, ``Toward characteristic-preserving image-based virtual try-on network,'' in \emph{Proceedings of the European Conference on Computer Vision (ECCV)}, 2018, pp. 589--604.

\bibitem{issenhuth2020parserfree}
T.~Issenhuth, J.~Mary, and C.~Calauzenes, ``Do not mask what you do not need to mask: A parser-free virtual try-on,'' in \emph{Proceedings of the European Conference on Computer Vision (ECCV)}, 2020, pp. 619--635.

\bibitem{ren2021clothtransformer}
B.~Ren, H.~Tang, F.~Meng \emph{et~al.}, ``Cloth interactive transformer for virtual try-on,'' \emph{arXiv preprint arXiv:2104.05519}, 2021.

\bibitem{xie2021patchgan}
Z.~Xie, Z.~Huang, F.~Zhao \emph{et~al.}, ``Towards scalable unpaired virtual try-on via patch-routed spatially-adaptive gan,'' in \emph{Advances in Neural Information Processing Systems (NeurIPS)}, 2021, pp. 2598--2610.

\bibitem{densepose}
R.~A. G{\"u}ler, N.~Neverova, and I.~Kokkinos, ``Densepose: Dense human pose estimation in the wild,'' in \emph{Proceedings of the IEEE Conference on Computer Vision and Pattern Recognition}, 2018, pp. 7297--7306.

\bibitem{openpose1}
Z.~{Cao}, G.~{Hidalgo Martinez}, T.~{Simon}, S.~{Wei}, and Y.~A. {Sheikh}, ``Openpose: Realtime multi-person 2d pose estimation using part affinity fields,'' \emph{IEEE Transactions on Pattern Analysis and Machine Intelligence}, 2019.

\bibitem{openpose2}
T.~Simon, H.~Joo, I.~Matthews, and Y.~Sheikh, ``Hand keypoint detection in single images using multiview bootstrapping,'' in \emph{CVPR}, 2017.

\bibitem{openpose3}
Z.~Cao, T.~Simon, S.-E. Wei, and Y.~Sheikh, ``Realtime multi-person 2d pose estimation using part affinity fields,'' in \emph{CVPR}, 2017.

\bibitem{openpose4}
S.-E. Wei, V.~Ramakrishna, T.~Kanade, and Y.~Sheikh, ``Convolutional pose machines,'' in \emph{CVPR}, 2016.

\bibitem{LIP}
K.~Gong, X.~Liang, D.~Zhang, X.~Shen, and L.~Lin, ``Look into person: Self-supervised structure-sensitive learning and a new benchmark for human parsing,'' in \emph{2017 IEEE Conference on Computer Vision and Pattern Recognition (CVPR)}, 2017, pp. 6757--6765.

\bibitem{fastddpm}
H.~Jiang, M.~Imran, L.~Ma, T.~Zhang, Y.~Zhou, M.~Liang, K.~Gong, and W.~Shao, ``Fast-ddpm: Fast denoising diffusion probabilistic models for medical image-to-image generation,'' \emph{arXiv preprint arXiv:2405.14802}, 2024.

\bibitem{RePaint}
\BIBentryALTinterwordspacing
A.~Lugmayr, M.~Danelljan, A.~Romero, F.~Yu, R.~Timofte, and L.~V. Gool, ``Repaint: Inpainting using denoising diffusion probabilistic models,'' \emph{2022 IEEE/CVF Conference on Computer Vision and Pattern Recognition (CVPR)}, pp. 11\,451--11\,461, 2022. [Online]. Available: \url{https://api.semanticscholar.org/CorpusID:246240274}
\BIBentrySTDinterwordspacing

\bibitem{FID}
\BIBentryALTinterwordspacing
M.~Heusel, H.~Ramsauer, T.~Unterthiner, B.~Nessler, G.~Klambauer, and S.~Hochreiter, ``Gans trained by a two time-scale update rule converge to a nash equilibrium,'' \emph{ArXiv}, vol. abs/1706.08500, 2017. [Online]. Available: \url{https://api.semanticscholar.org/CorpusID:231697514}
\BIBentrySTDinterwordspacing

\bibitem{LPIPS}
\BIBentryALTinterwordspacing
R.~Zhang, P.~Isola, A.~A. Efros, E.~Shechtman, and O.~Wang, ``The unreasonable effectiveness of deep features as a perceptual metric,'' \emph{2018 IEEE/CVF Conference on Computer Vision and Pattern Recognition}, pp. 586--595, 2018. [Online]. Available: \url{https://api.semanticscholar.org/CorpusID:4766599}
\BIBentrySTDinterwordspacing

\bibitem{PaintbyExample}
\BIBentryALTinterwordspacing
B.~Yang, S.~Gu, B.~Zhang, T.~Zhang, X.~Chen, X.~Sun, D.~Chen, and F.~Wen, ``Paint by example: Exemplar-based image editing with diffusion models,'' \emph{2023 IEEE/CVF Conference on Computer Vision and Pattern Recognition (CVPR)}, pp. 18\,381--18\,391, 2022. [Online]. Available: \url{https://api.semanticscholar.org/CorpusID:253802085}
\BIBentrySTDinterwordspacing

\bibitem{GP-VTON}
\BIBentryALTinterwordspacing
Z.~Xie, Z.~Huang, X.~Dong, F.~Zhao, H.~Dong, X.~Zhang, F.~Zhu, and X.~Liang, ``Gp-vton: Towards general purpose virtual try-on via collaborative local-flow global-parsing learning,'' \emph{2023 IEEE/CVF Conference on Computer Vision and Pattern Recognition (CVPR)}, pp. 23\,550--23\,559, 2023. [Online]. Available: \url{https://api.semanticscholar.org/CorpusID:257757040}
\BIBentrySTDinterwordspacing

\end{thebibliography}




\end{document}